\documentclass[runningheads]{llncs}
\usepackage[normalem]{ulem}
\usepackage[T1]{fontenc}
\usepackage[nocompress]{cite}
\usepackage{amsfonts}
\usepackage{multicol}
\usepackage{multirow}
\usepackage{booktabs}
\usepackage{array}
\usepackage{arydshln}
\usepackage{xcolor}
\usepackage{colortbl}
\usepackage{algorithmicx}
\usepackage{subcaption}
\usepackage{algorithm}
\usepackage[noend]{algpseudocode}
\usepackage{amsmath}
\usepackage{amssymb}
\usepackage{tcolorbox}

\usepackage{graphicx,verbatim}
\usepackage[misc]{ifsym}

\begin{document}
\newcolumntype{P}[1]{>{\centering\arraybackslash}p{#1}}
\titlerunning{MergeSurv}
\title{MergeSurv: Merging-Based Continual Learning for Survival Analysis on Whole-Slide Images}
%

\author{Vu Minh Tran$^{1,2}$, Doanh C. Bui$^{1,2}$, Van-Hau Pham$^{1,2}$, Khang Nguyen$^{1,2}$}  
\authorrunning{Tran et al.}
\institute{$^1$University of Information Technology \\ $^2$Viet Nam National University Ho Chi Minh City}

    
\maketitle     
\begin{abstract}

Survival analysis on Whole Slide Images (WSIs) is important in computational pathology for prognosis estimation and treatment planning. However, existing survival models are typically trained independently for each cancer cohort, making continual adaptation computationally expensive for gigapixel-scale WSIs. In this study, we propose MergeSurv, a merging-based continual learning framework for WSI survival analysis. A pathology vision-language foundation model is independently fine-tuned on each task, and the learned parameters are sequentially merged into a unified model without storing previous training data. We further investigate two inference strategies: One-for-All (OFA) and Voting-Expert Aggregation (VEA). 
Experiments on four TCGA cohorts demonstrate that MergeSurv provides a favorable trade-off between survival prediction performance, forgetting mitigation, and memory-free continual learning. The results suggest that model merging is a promising direction for scalable and privacy-preserving continual learning in computational pathology.


\end{abstract}

\section{Introduction}

Survival risk prediction is an important task in computational pathology because it supports prognosis estimation and treatment planning for cancer patients \cite{cox1972regression,yao2020deepattnmisl}. Whole Slide Images (WSIs), which provide detailed tissue information at the cellular level, are widely used for this purpose \cite{clam2021,yao2020deepattnmisl}. However, existing survival prediction models are still largely task-specific, where a new standalone model must be trained whenever a new organ or cohort is introduced \cite{yao2020deepattnmisl,yu2026consurv}. This limitation is particularly problematic for WSI analysis because WSIs are gigapixel-scale images requiring substantial storage, preprocessing, and training resources \cite{clam2021}. Therefore, continual learning (CL), which enables models to learn sequential tasks while mitigating catastrophic forgetting \cite{kirkpatrick2017overcoming,wang2024continualsurvey}, has become a promising direction for WSI analysis.

Existing CL approaches mainly include regularization-based methods, such as EWC \cite{kirkpatrick2017overcoming} and LwF \cite{li2018learning}, and rehearsal-based methods, such as ER \cite{rolnick2019experience}, DER, and DER++ \cite{buzzega2020der}. However, rehearsal-based methods require storing previous WSIs, leading to high memory costs and potential privacy concerns in medical applications. Moreover, most existing WSI CL studies focus on classification tasks \cite{bui2026merge,lee2025comel,huang2023conslide}, while continual survival prediction remains underexplored \cite{yu2026consurv}. Recent studies have explored model merging \cite{wortsman2022modelsoups,ilharco2023taskarithmetic}, which integrates independently trained task-specific models into a unified model without storing original training data. This property makes model merging particularly suitable for lifelong WSI learning, where repeated training and cross-institutional data sharing are often expensive. 
\textit{Although MergeSlide \cite{bui2026merge} recently introduced merging-based continual learning for WSI classification, it does not address survival-specific challenges such as censored data and continuous risk prediction.}

In this study, we propose MergeSurv, a merging-based continual learning framework for survival analysis on WSIs. Specifically, a pre-trained pathology foundation model (TITAN \cite{ding2025multimodal}) is independently fine-tuned on each survival task to obtain task-specific parameters, which are subsequently merged into a unified model. During inference, we investigate two prediction strategies: \textbf{1) One-for-All (OFA)}, which uses a single unified survival head, and \textbf{2) Voting-Expert Aggregation (VEA)}, which combines predictions from multiple task-specific heads through text-prompt-based routing.

\noindent \textbf{The main contributions are summarized as follows:}
\begin{itemize}
    \item We propose \textbf{MergeSurv}, a merging-based continual
    learning framework tailored to WSI-based survival analysis with censored
    outcomes, enabling sequential adaptation without retaining data from
    previous tasks.
    \item We introduce \textbf{OFA} and \textbf{VEA} for survival risk prediction
    with a merged WSI aggregator. In particular, VEA enables
    task-identity-free routing among task-specific Cox experts using pathology
    vision-language representations.
    \item We conduct an empirical comparison with representative
    regularization- and rehearsal-based continual learning methods on four
    TCGA cohorts using both standard and censoring-adjusted survival metrics.
\end{itemize}

\section{Related Work}

\noindent\textbf{Continual Learning.} Continual Learning (CL) trains models on sequential tasks while mitigating catastrophic forgetting \cite{kirkpatrick2017overcoming,wang2024continualsurvey}. Existing approaches mainly include regularization-based methods, such as EWC \cite{kirkpatrick2017overcoming} and LwF \cite{li2018learning}, and rehearsal-based methods, such as ER \cite{rolnick2019experience}, DER, and DER++ \cite{buzzega2020der}. However, rehearsal-based methods require storing previous data, increasing memory cost and privacy concerns in medical applications. In computational pathology, CL is important because Whole Slide Images (WSIs) are gigapixel-scale data with high computational and storage demands \cite{clam2021}. Existing WSI CL studies \cite{bui2026merge,bui2025zeroslide,lee2025comel,huang2023conslide} mainly focus on classification, while survival prediction remains underexplored. MergeSlide \cite{bui2026merge} recently introduced merging-based continual learning for WSI classification, but it does not address survival-specific challenges such as censored data and continuous risk scores.

\noindent\textbf{Model Merging.} Model merging integrates independently trained models into a unified model while preserving task-specific knowledge \cite{wortsman2022modelsoups,ilharco2023taskarithmetic}. Most methods represent each task as a task vector, defined as the parameter difference between a task-specific model and a shared base model, and combine them through different merging strategies \cite{ilharco2023taskarithmetic,tang2025merging}. Recent studies further explore weighted merging, Mixture-of-Experts routing, and orthogonality constraints to reduce task conflicts \cite{wortsman2022modelsoups,tang2025merging}. Since model merging avoids storing original training data, it is well suited for lifelong WSI learning where data sharing and retraining are costly.

\section{Methodology}

\subsection{An Overview of MergeSurv}

\begin{figure}[t]
    \centering
    \includegraphics[width=0.83\linewidth]{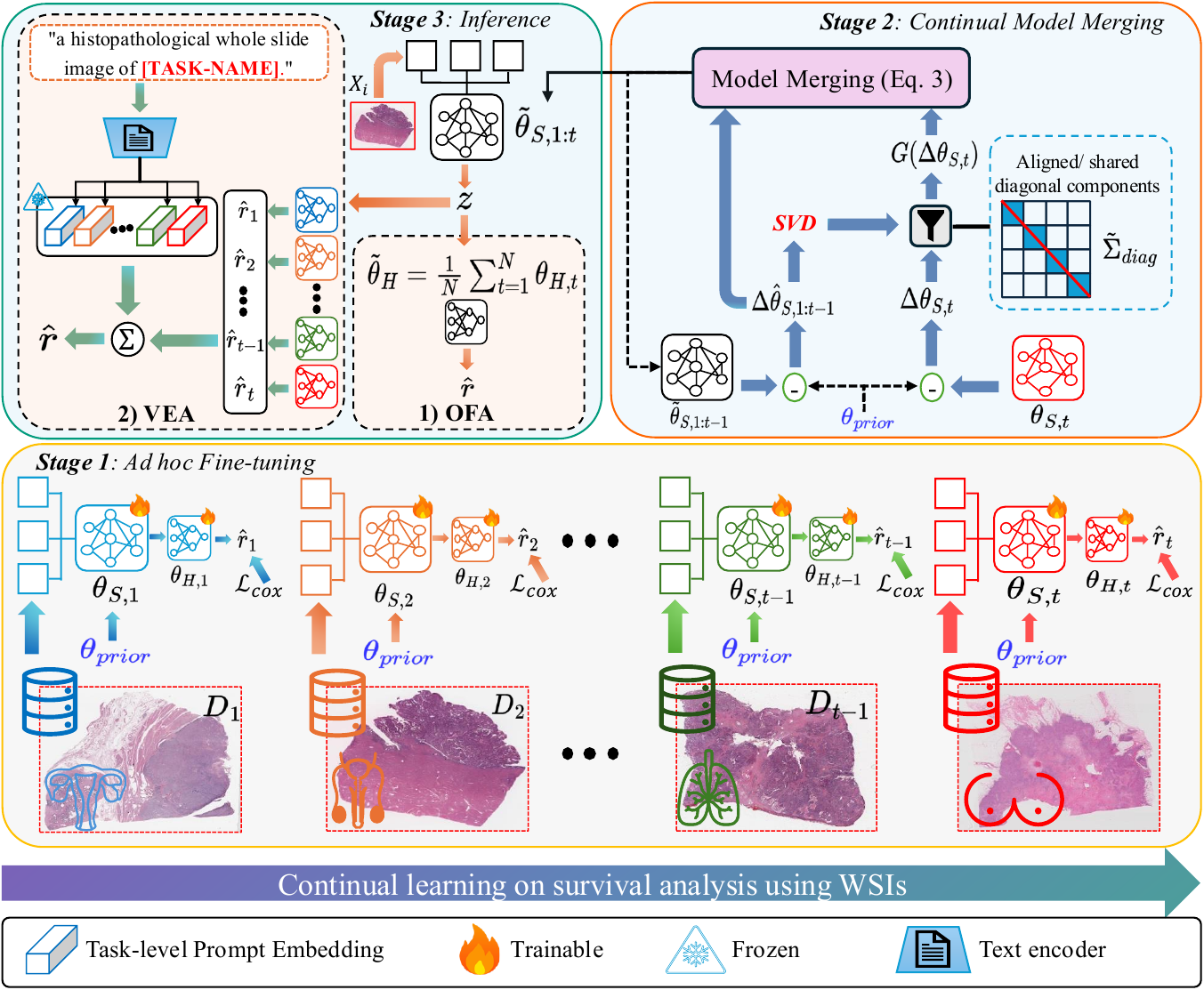}
    \caption{Overall framework of MergeSurv. Each survival task is fine-tuned from TITAN pretrained weights $\theta_{prior}$ using the WSI model $f_{\mathcal{S}}$ with a Cox regressor. Task-specific WSI aggregator parameters are sequentially merged using Eq.~\ref{eq:merge}. Two inference strategies are introduced: One-for-All (OFA) with a single Cox head, and Voting-Expert Aggregation (VEA) with multiple task-specific heads.}
    \label{fig:overall_architecture}
\end{figure}

\textbf{MergeSurv} is a CL framework for survival analysis using WSIs. Given a stream of $N$ datasets $\mathcal{D}=\{D_t\}_{t=1}^{N}$ corresponding to $N$ tasks, MergeSurv trains each survival task $D_t$ using the WSI model $f_\mathcal{S}(\cdot;\theta_\mathcal{S})$ with a Cox regression head $H(\cdot;\theta_H)$ to obtain the task-specific optimal parameters $\theta^*_{\mathcal{S},t}$ and $\theta^*_{H,t}$. Subsequently, all parameters $\{\theta^*_{\mathcal{S},t}\}_{t=1}^{N}$ are sequentially merged into unified parameters, resulting in the merged parameters $\tilde{\Theta}_{\mathcal{S},1:N}$. Finally, only $f_\mathcal{S}(\cdot;\tilde{\Theta}_{\mathcal{S},1:N})$ and two risk prediction strategies to estimate survival risk scores are used for inference on all $N$ tasks learned so far, as illustrated in Fig.~\ref{fig:overall_architecture}.

\subsection{\textit{Ad Hoc} Fine-Tuning}

\subsubsection{Fine-tuning Process}
First, whenever a new WSI survival task $t$ is defined, the model $f_{\mathcal{S}}$ is fine-tuned from the prior parameters $\theta_{\text{prior}}$ using the training samples of $D_t$ to optimize the objective function $\mathcal{L}$ over $E$ epochs. $\theta_{\text{prior}}$ is initialized from the pretrained weights of the TITAN foundation model \cite{ding2025multimodal}. As TITAN is a pathology-specific foundation model trained on large-scale unlabeled pathology images in a self-supervised manner, it effectively captures tissue structures in the embedding space. The fine-tuning process is formulated as follows.

\begin{equation}
\begin{aligned}
\{\theta_{\mathcal{S},t}^*,\theta_{H,t}^*\}
&
\approx
\arg\min_{\theta_{\mathcal{S},t},\theta_{H,t}}
\mathcal{L}
\left(
\hat r_i, Y_i
\right),
\quad
(X_i,Y_i)\sim D_t,
\quad
\theta_{S,t}^{(0)}=\theta_{\text{prior}},
\quad
e=1\rightarrow E, \\
&\text{where }
\quad
Y_i=(T_i,\delta_i), 
\quad
\hat{r_i} = H(f_{\mathcal{S}}(X_i;\theta_{\mathcal{S},t});\theta_{H,t})
\end{aligned}
\label{eq:train}
\end{equation}

\noindent where $\{\theta_{\mathcal{S},t}^*,\theta_{H,t}^*\}$ denotes the optimal parameters of the WSI model and the regression head for task $t$ and $X_i = \{x_k\}_{k=1}^{K}$ denotes a bag of $K$ patch embeddings extracted from an input WSI, where $x_k \in \mathbb{R}^{1\times d_{\text{vis}}}$. $H$ is a Cox regression head \cite{cox1972regression,yao2020deepattnmisl} that predicts a scalar risk score $\hat{r_i} \in \mathbb{R}$, where a larger value indicates a higher event risk. Therefore, the fine-tuning stage jointly learns the WSI aggregator and the task-specific Cox head by optimizing the loss function.

\subsubsection{Loss Function} 

We optimize the model using the Cox partial likelihood loss $\mathcal{L}_{Cox}$, which is defined as follows:

\begin{equation}
\mathcal{L}_{{Cox}} =
-\frac{1}{N_e}
\sum_{i:\delta_i=1}
\left(
\hat{r_i} -
\log \sum_{j:T_j \geq T_i} \exp(\hat{r_j})
\right),
\end{equation}

\noindent where $N_e=\sum_{i} \delta_i$ is the number of uncensored event samples, $T_i$ denotes the survival time, and $\delta_i$ is the event indicator, with $\delta_i=1$ indicating death from any cause and $\delta_i=0$ indicating right censoring at the last follow-up or loss to follow-up. The predicted risk score for the $i$-th sample is denoted by $\hat{r}_i$. The Cox loss learns the relative risk ranking among patients, assigning higher risk scores to patients with earlier events.

\subsection{Model Merging Operation}

We follow OPCM \cite{tang2025merging} for the model merging operations of the WSI model parameters $f_{\mathcal{S}}$. Three sets of parameters are considered: 1) the prior parameters $\theta_{\mathrm{prior}} \in \mathbb{R}^{m\times n}$, 2) the previously merged parameters of the WSI model $\tilde{\Theta}_{\mathcal{S},1:t-1} \in \mathbb{R}^{m\times n}$, and 3) the newly fine-tuned parameters $\theta^*_{\mathcal{S},t} \in \mathbb{R}^{m\times n}$. The merged parameters are updated as follows:

\begin{equation}
\tilde{\Theta}_{\mathcal{S},1:t}
=
\theta_{\mathrm{prior}}
+
\frac{\lambda_{t-1}\Delta\widetilde{\theta}_{\mathcal{S},1:t-1} + G(\Delta\theta_{\mathcal{S},t}, \Delta\widetilde{\theta}_{\mathcal{S},1:t-1})}{\lambda_t},
\label{eq:merge}
\end{equation}

\noindent where $\Delta\theta_{\mathcal{S},t}=\theta^*_{\mathcal{S},t}-\theta_{\mathrm{prior}}$ and $\Delta\widetilde{\theta}_{\mathcal{S},1:t-1}=\tilde{\Theta}_{\mathcal{S},1:t-1}-\theta_{\mathrm{prior}}$ denote the current and accumulated task vectors, respectively  \cite{ilharco2023taskarithmetic}. $G(\cdot)$ is a projection function, and $\lambda_t$ \cite{bui2026merge,tang2025merging} controls the balance between old and new knowledge. To preserve task separability, OPCM projects the new task vector $\Delta\theta_{\mathcal{S},t}$ onto the orthogonal space of $\Delta\widetilde{\theta}_{\mathcal{S},1:t-1}$. Specifically, singular value decomposition (SVD) is applied to $\Delta\widetilde{\theta}_{\mathcal{S},1:t-1}$ to obtain orthogonal subspaces, and the components of $\Delta\theta_{\mathcal{S},t}$ aligned with previous task directions are removed through a zero-diagonal masking operation. This projection satisfies $\langle G(\Delta\theta_{\mathcal{S},t}), \Delta\widetilde{\theta}_{\mathcal{S},1:t-1} \rangle_F = 0$, ensuring that the newly merged knowledge does not interfere with previously learned tasks.

To further prevent the merged model from drifting excessively from the base model, the balancing factor $\lambda_t$ is adaptively computed based on the norm ratio between the accumulated merged task vectors and the average norm of all task-specific parameters. As a result, the distance between the merged model and the base model remains bounded by the average task norm, stabilizing the continual merging process across sequential survival tasks.

\subsection{Risk Prediction}

\noindent\textbf{One-for-All.}
In the one-for-all (OFA) strategy, the WSI aggregator is merged using OPCM to obtain the merged parameters $\tilde{\Theta}_{\mathcal{S},1:N}$, while the survival head is constructed by averaging the weights of the learned task-specific Cox heads. Let \(\theta_{H,t}^{*}\) denote the parameters of the Cox head \(H\) learned for task \(t\). After observing \(N\) tasks, the merged Cox head parameters are computed by averaging all task-specific heads, and the final risk score is computed as:
\begin{equation}
\hat{r}
=
H(f_\mathcal{S}(X;\tilde{\Theta}_{\mathcal{S},1:N});\tilde{\theta}_H),
\quad
\text{where }
\tilde{\theta}_H
=
\frac{1}{N}\sum_{t=1}^{N}\theta_{H,t}^{*}.
\end{equation}

This strategy creates a unified model consisting of one merged backbone and a single Cox head; therefore, \textit{OFA is simple and memory-efficient during inference.}

\begin{table}[!t]
\centering
\caption{Task-level prompts used for survival cohort routing.}
\label{tab:survival_prompts}
\subcaption{\textbf{Task-level prompt template variations}}
\begin{tabular}{cl}
\toprule
\# & \textbf{Prompt template} \\
\midrule

\textcircled{\scriptsize 1}
& ``\texttt{[TASKNAME].}'' \\

\textcircled{\scriptsize 2}
& ``\texttt{a histopathological whole slide image of [TASKNAME].}'' \\

\textcircled{\scriptsize 3}
& ``\texttt{an H\&E stained whole slide image of [TASKNAME].}'' \\

\textcircled{\scriptsize 4}
& ``\texttt{a pathology image from [TASKNAME].}'' \\

\bottomrule
\end{tabular}

\vspace{-0.2cm}

\subcaption{\textbf{\texttt{[TASKNAME]} entries used for each survival cohort}}
\begin{tabular}{cl}
\toprule
\textbf{Cohort} & \textbf{\texttt{[TASKNAME]}} \\
\midrule

\multirow{3}{*}{BLCA}
& \textcircled{\scriptsize 1} ``bladder urothelial carcinoma'' \\
& \textcircled{\scriptsize 2} ``bladder cancer'' \\
& \textcircled{\scriptsize 3} ``TCGA-BLCA'' \\
\midrule

\multirow{3}{*}{UCEC}
& \textcircled{\scriptsize 1} ``uterine corpus endometrial carcinoma'' \\
& \textcircled{\scriptsize 2} ``endometrial cancer'' \\
& \textcircled{\scriptsize 3} ``TCGA-UCEC'' \\
\midrule

\multirow{3}{*}{LUAD}
& \textcircled{\scriptsize 1} ``lung adenocarcinoma'' \\
& \textcircled{\scriptsize 2} ``lung cancer adenocarcinoma'' \\
& \textcircled{\scriptsize 3} ``TCGA-LUAD'' \\
\midrule

\multirow{3}{*}{BRCA}
& \textcircled{\scriptsize 1} ``breast invasive carcinoma'' \\
& \textcircled{\scriptsize 2} ``breast cancer'' \\
& \textcircled{\scriptsize 3} ``TCGA-BRCA'' \\

\bottomrule
\end{tabular}

\end{table}

\noindent\textbf{Voting-Expert Aggregation.}
In the voting-expert aggregation (VEA) strategy, the WSI aggregator is still merged using OPCM, while the task-specific Cox head parameters \(\{\theta^*_{H,t}\}_{t=1}^{N}\) are retained instead of being merged into a single head. In this manner, each head acts as an expert and produces an individual risk score:
\begin{equation}
\hat{r}_t = H(f_{\mathcal{S}}(X;\tilde{\Theta}_{\mathcal{S},1:N});\theta_{H,t}^{*}),
\quad t=1,\ldots,N.
\end{equation}

When the task identity is not provided during inference, MergeSurv estimates the relevance of each expert using task-level text prompts.
For each survival cohort, we consider three task-name variants, including the full cancer name, a common alias, and the corresponding TCGA cohort code. These task names are combined with four pathology-oriented prompt templates to construct the cohort-specific prompt set. The prompt templates and task-name entries are shown in Table~\ref{tab:survival_prompts}.

The generated prompts are encoded using the frozen TITAN text encoder to
derive the task-level representation.
Let \(E_{\text{task},t}\) denote the text embedding of the task-level prompt corresponding to task \(t\). The routing weight is computed using the cosine similarity between the WSI representation obtained by WSI model $f_\mathcal{S}$ and the task-level text representation:
$
\pi_t =
\frac{\exp(a_t/\gamma)}
{\sum_{j=1}^{N}\exp(a_j/\gamma)}
$, where
$a_t = \cos\big(f_{\mathcal{S}}(X;\tilde{\Theta}_{\mathcal{S},1:N}),E_{\text{task},t}\big)$.

The final risk score is computed as the weighted sum of the predictions from all expert heads:
\begin{equation}
\hat{r} = \sum_{t=1}^{N}\pi_t \hat{r}_t.
\end{equation}

This strategy helps preserve the task-specific knowledge of each survival head \textit{while still using a single merged WSI aggregator to extract WSI representations.}

\section{Experiments}

\begin{table}[!t]
\centering
\caption{Statistics of the four TCGA cohorts.}
\label{tab:dataset_statistics}
\begin{tabular}{cccccc}
\toprule
Dataset
& Cancer Type
& WSIs
& Cens. WSIs
& Patients
& Cens. Patients \\
\midrule
TCGA-BLCA & Bladder      & 437  & 236 & 373 & 204 \\
TCGA-UCEC & Endometrial  & 539  & 460 & 480 & 405 \\
TCGA-LUAD & Lung         & 516  & 314 & 453 & 295 \\
TCGA-BRCA & Breast       & 1022 & 888 & 956 & 826 \\
\bottomrule
\end{tabular}
\end{table}

\subsubsection{Datasets}
We use four cancer cohorts from The Cancer Genome Atlas Program (TCGA) to evaluate WSI-based continual survival prediction: Bladder Urothelial Carcinoma (BLCA), Uterine Corpus Endometrial Carcinoma (UCEC), Lung Adenocarcinoma (LUAD), and Breast Invasive Carcinoma (BRCA). The numbers of WSIs, patients, and censored cases in each cohort are summarized in Table~\ref{tab:dataset_statistics}. Each cohort is regarded as a separate task in the continual learning sequence. Patient-level overall survival time and censoring status are used to define the survival prediction targets. To evaluate MergeSurv and other continual learning methods, we perform 5-fold cross-validation using patient-level splits, ensuring that all WSIs from the same patient are assigned to the same fold. The task order used in this study is BLCA$\rightarrow$UCEC$\rightarrow$LUAD$\rightarrow$BRCA.

\subsubsection{Experimental Settings.}
We evaluate MergeSurv on a continual learning sequence of four TCGA cohorts: BLCA, UCEC, LUAD, and BRCA. \textit{Joint training} is included as an upper-bound setting, while \textit{Naive fine-tuning} serves as a sequential learning baseline without forgetting mitigation. We further compare with representative continual learning methods, including EWC \cite{kirkpatrick2017overcoming}, LwF \cite{li2018learning}, ER \cite{rolnick2019experience}, DER \cite{buzzega2020der}, and DER++ \cite{buzzega2020der}, where replay-based methods use a buffer size of 32 WSIs. All experiments are conducted on an NVIDIA A100 (80GB) GPU.

\subsubsection{Implementation Details.}
WSIs are tiled into $256 \times 256$ patches at $10\times$ magnification. Patch and prompt embeddings are extracted using the TITAN encoders, which were pretrained on a large-scale pathology corpus excluding TCGA data~\cite{ding2025multimodal}. Therefore, the four TCGA cohorts used in our experiments do not overlap with the TITAN pretraining set. Following the MIL setting~\cite{clam2021}, each WSI is represented as a bag of instances with $K=1000$ randomly sampled patches. MergeSurv employs a Transformer-based WSI aggregator $f_\mathcal{S}$ initialized from TITAN pretrained weights $\theta_{prior}$. All models are trained for $E=20$ epochs using AdamW with learning rates of $1\times10^{-5}$ for the backbone and $1\times10^{-4}$ for the head.

\subsubsection{Main Results.}
We use the C-Index \cite{harrell1996multivariable} and C-Index IPCW \cite{uno2011cstatistics} to evaluate survival risk ranking, while Forgetting \cite{chaudhry2018riemannian} and Backward Transfer (BWT) \cite{lopez2017gradient} are used for continual learning evaluation. As shown in Table~\ref{tab:survival_results}, MergeSurv achieves strong continual survival prediction performance while effectively mitigating catastrophic forgetting across sequential TCGA cohorts.
For the standard C-Index in Table~\ref{tab:survival_results}(a), MergeSurv (VEA) achieves the best average C-Index ($0.6773$) and the lowest forgetting ($0.0138$), while MergeSurv (OFA) obtains the best BWT ($-0.0054$). Notably, MergeSurv (VEA) slightly outperforms Joint training ($0.6773$ vs.~$0.6739$), suggesting that parameter merging effectively preserves and integrates survival knowledge without joint training. Compared with replay-based methods, MergeSurv substantially reduces forgetting without requiring memory buffers.
For the IPCW-adjusted C-Index in Table~\ref{tab:survival_results}(b), EWC achieves the highest average score ($0.6353$), while MergeSurv (VEA) obtains the second-best score ($0.6294$) together with the lowest forgetting ($0.0159$) and competitive BWT ($0.0011$). 

\begin{table}[!t]
\centering
\caption{Continual survival prediction results on four TCGA cohorts. Best results are highlighted in \textbf{bold}, and the second-best results are \underline{underlined}. }
\label{tab:survival_results}
\subcaption{\textbf{C-Index}}
\resizebox{\textwidth}{!}{%
\begin{tabular}{lcccc}
\toprule
\textbf{Method} 
& \textbf{Buffer} 
& \textbf{Average ($\uparrow$)} 
& \textbf{Forget ($\downarrow$)} 
& \textbf{BWT ($\uparrow$)} \\
\midrule

Joint training
& -- 
& 0.6739 ($\pm$0.0125) 
& -- 
& -- \\

Naive fine-tuning 
& -- 
& 0.6618 ($\pm$0.0274) 
& 0.0303 ($\pm$0.0133) 
& -0.0161 ($\pm$0.0260) \\ 
\midrule

EWC 
& -- 
& 0.6596 ($\pm$0.0326) 
& 0.0489 ($\pm$0.0449) 
& -0.0363 ($\pm$0.0590) \\

LwF 
& -- 
& 0.6531 ($\pm$0.0276) 
& \uline{0.0236 ($\pm$0.0196)} 
& -0.0123 ($\pm$0.0255) \\

ER 
& 32 WSIs 
& 0.6535 ($\pm$0.0289) 
& 0.0549 ($\pm$0.0217) 
& -0.0468 ($\pm$0.0194) \\

DER 
& 32 WSIs 
& 0.6630 ($\pm$0.0259) 
& 0.0386 ($\pm$0.0266) 
& -0.0196 ($\pm$0.0356) \\

DER++ 
& 32 WSIs 
& 0.6620 ($\pm$0.0334) 
& 0.0500 ($\pm$0.0396) 
& -0.0445 ($\pm$0.0444) \\ 
\midrule

MergeSurv (OFA) 
& -- 
& \uline{0.6693 ($\pm$0.0245)}
& 0.0262 ($\pm$0.0167) 
& \textbf{-0.0054 ($\pm$0.0227)} \\

MergeSurv (VEA)
& -- 
& \textbf{0.6773 ($\pm$0.0133)} 
& \textbf{0.0138 ($\pm$0.0082)} 
& \uline{-0.0097 ($\pm$0.0115)} \\

\bottomrule
\end{tabular}
}

\vspace{-0.2cm}

\subcaption{\textbf{C-Index IPCW}}
\resizebox{\textwidth}{!}{%
\begin{tabular}{lcccc}
\toprule
\textbf{Method} 
& \textbf{Buffer}
& \textbf{Average ($\uparrow$)} 
& \textbf{Forget ($\downarrow$)} 
& \textbf{BWT ($\uparrow$)} \\
\midrule

Joint training
& -- 
& 0.6184 ($\pm$0.0386) 
& -- 
& -- \\

Naive fine-tuning 
& -- 
& 0.6099 ($\pm$0.0621) 
& 0.0282 ($\pm$0.0210) 
& -0.0068 ($\pm$0.0339) \\ 
\midrule

EWC 
& -- 
& \textbf{0.6353 ($\pm$0.0474)} 
& 0.0411 ($\pm$0.0150) 
& \uline{0.0040 ($\pm$0.0754)}\\

LwF 
& -- 
& 0.6045 ($\pm$0.0663) 
& \uline{0.0264 ($\pm$0.0273)} 
& \textbf{0.0164 ($\pm$0.0745)} \\

ER 
& 32 WSIs 
& 0.6082 ($\pm$0.0477) 
& 0.0570 ($\pm$0.0454) 
& -0.0072 ($\pm$0.0452) \\

DER 
& 32 WSIs 
& 0.6246 ($\pm$0.0771) 
& 0.0300 ($\pm$0.0116) 
& -0.0056 ($\pm$0.0201) \\

DER++ 
& 32 WSIs 
& 0.6014 ($\pm$0.0698) 
& 0.0511 ($\pm$0.0413) 
& -0.0288 ($\pm$0.0419) \\ 
\midrule

MergeSurv (OFA) 
& -- 
& 0.6133 ($\pm$0.0571) 
& 0.0639 ($\pm$0.0718) 
& -0.0275 ($\pm$0.0678) \\

MergeSurv (VEA)
& -- 
& \uline{0.6294 ($\pm$0.0497)} 
& \textbf{0.0159 ($\pm$0.0099)} 
& 0.0011 ($\pm$0.0346) \\

\bottomrule
\end{tabular}
}
\end{table}

\section{Additional Analysis}
\begin{table}[!t]
\centering
\caption{Cohort-wise C-Index of MergeSurv-VEA across continual learning steps. Rows denote the most recently learned cohort, and columns denote the evaluation cohort.}
\label{tab:vea_cohort_matrix}
\resizebox{\textwidth}{!}{%
\begin{tabular}{ccccc}
\toprule
\textbf{Step}
& \textbf{BLCA}
& \textbf{UCEC}
& \textbf{LUAD}
& \textbf{BRCA} \\
\midrule
\textbf{BLCA}
& 0.6694 ($\pm$0.0554)
& --
& --
& -- \\

\textbf{UCEC}
& 0.6595 ($\pm$0.0610)
& 0.7457 ($\pm$0.0329)
& --
& -- \\

\textbf{LUAD}
& 0.6620 ($\pm$0.0643)
& 0.7495 ($\pm$0.0268)
& 0.6237 ($\pm$0.0746)
& -- \\

\textbf{BRCA}
& 0.6587 ($\pm$0.0676)
& 0.7328 ($\pm$0.0340)
& 0.6181 ($\pm$0.0755)
& 0.6996 ($\pm$0.0149) \\
\bottomrule
\end{tabular}
}
\end{table}

\subsection{Cohort-wise Continual Performance}
To provide a more fine-grained view of performance across individual cancer cohorts, Table~\ref{tab:vea_cohort_matrix} reports the cohort-wise C-Index of MergeSurv-VEA after each continual learning step. This analysis complements the aggregate metrics by showing how performance on previously learned cohorts
changes as new tasks are introduced.

After completing the full sequence, MergeSurv-VEA achieves C-Indices of 0.6587, 0.7328, 0.6181, and 0.6996 on BLCA, UCEC, LUAD, and BRCA, respectively. Compared with the performance obtained immediately after learning each cohort, the final C-Index decreases by 0.0107 on BLCA, 0.0129 on UCEC, and 0.0056 on LUAD. These relatively small changes are consistent with the low forgetting reported in Table~\ref{tab:survival_results}. Among the four cohorts, UCEC achieves the highest final C-Index, whereas LUAD remains the most challenging cohort under the evaluated task sequence.

\subsection{Effects of Task Sequence}

\begin{table}[b]
\centering
\caption{Performance comparison of MergeSurv-VEA across various continual learning sequences, denoted as follows: \textcircled{\scriptsize 1} BLCA $\rightarrow$ UCEC $\rightarrow$ LUAD $\rightarrow$ BRCA; \textcircled{\scriptsize 2} BRCA $\rightarrow$ LUAD $\rightarrow$ UCEC $\rightarrow$ BLCA; \textcircled{\scriptsize 3} LUAD $\rightarrow$ BLCA $\rightarrow$ UCEC $\rightarrow$ BRCA.}
\label{tab:order_performance}
\resizebox{\textwidth}{!}{%
\begin{tabular}{ccccccc}
\toprule
\multirow{2}{*}{\textbf{Order}} 
& \multicolumn{3}{c}{\textbf{C-Index}} 
& \multicolumn{3}{c}{\textbf{C-Index IPCW}} \\
\cmidrule(lr){2-4} \cmidrule(lr){5-7}
& \textbf{Average ($\uparrow$)} 
& \textbf{Forget ($\downarrow$)} 
& \textbf{BWT ($\uparrow$)} 
& \textbf{Average ($\uparrow$)} 
& \textbf{Forget ($\downarrow$)} 
& \textbf{BWT ($\uparrow$)} \\
\midrule

\multirow{2}{*}{\textcircled{\scriptsize 1}}
& \textbf{0.6773} & 0.0138 & -0.0097 & \textbf{0.6294} & 0.0159 & \textbf{0.0011} \\
& ($\pm$0.0133) & ($\pm$0.0082) & ($\pm$0.0115) & ($\pm$0.0497) & ($\pm$0.0099) & ($\pm$0.0346) \\
\addlinespace[2pt]

\multirow{2}{*}{\textcircled{\scriptsize 2}}
& 0.6772 & \textbf{0.0075} & \textbf{-0.0027} & 0.6170 & \textbf{0.0094} & -0.0025 \\
& ($\pm$0.0138) & ($\pm$0.0033) & ($\pm$0.0066) & ($\pm$0.0454) & ($\pm$0.0031) & ($\pm$0.0042) \\
\addlinespace[2pt]

\multirow{2}{*}{\textcircled{\scriptsize 3}}
& 0.6772 & 0.0125 & -0.0107 & 0.6182 & 0.0190 & -0.0162 \\
& ($\pm$0.0136) & ($\pm$0.0053) & ($\pm$0.0041) & ($\pm$0.0455) & ($\pm$0.0136) & ($\pm$0.0158) \\
\bottomrule
\end{tabular}%
}
\end{table}

To investigate the effect of task order on the proposed method, we evaluate MergeSurv-VEA under three different continual learning sequences. As shown in Table~\ref{tab:order_performance}, the method is relatively insensitive to changes in task order. In particular, the average C-Index remains nearly unchanged across all sequences, ranging only from 0.6772 to 0.6773, indicating stable overall risk discrimination. Forgetting and BWT also exhibit only minor variations, although each sequence provides a slight advantage for a particular metric. 

\par
\clearpage
\begin{figure}[!p]
\centering
\setlength{\tabcolsep}{2pt}

\begin{tabular}{cccc}

\textbf{BLCA} & \textbf{UCEC} & \textbf{LUAD} & \textbf{BRCA} \\[0.1cm]

\begin{subfigure}{0.24\textwidth}
    \centering
    \includegraphics[width=\linewidth]{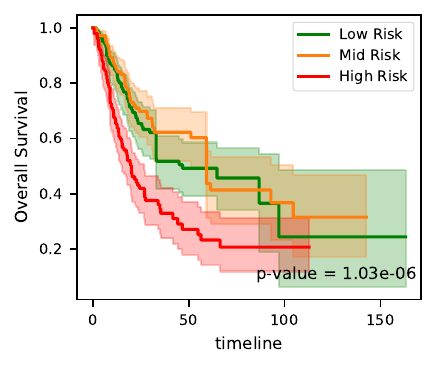}
\end{subfigure}
&
\begin{subfigure}{0.24\textwidth}
    \centering
    \includegraphics[width=\linewidth]{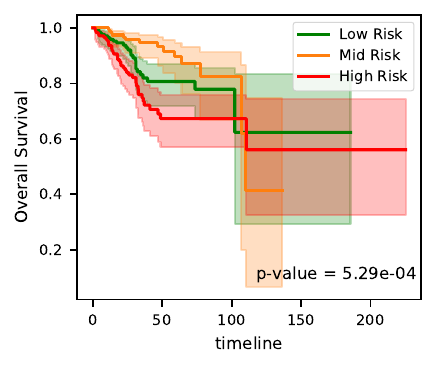}
\end{subfigure}
&
\begin{subfigure}{0.24\textwidth}
    \centering
    \includegraphics[width=\linewidth]{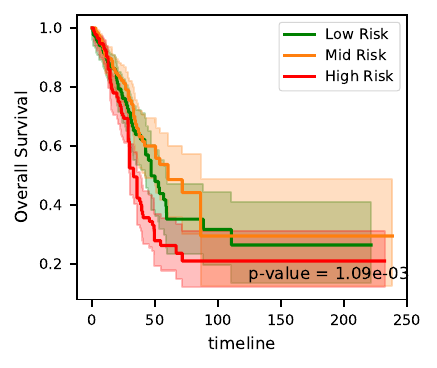}
\end{subfigure}
&
\begin{subfigure}{0.24\textwidth}
    \centering
    \includegraphics[width=\linewidth]{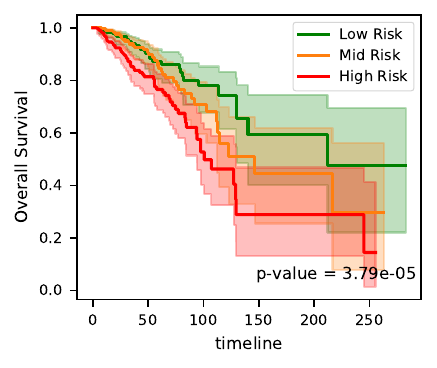}
\end{subfigure}
\\[-0.05cm]

\multicolumn{4}{c}{\textbf{(a) EWC}}
\\[0.35cm]

\begin{subfigure}{0.24\textwidth}
    \centering
    \includegraphics[width=\linewidth]{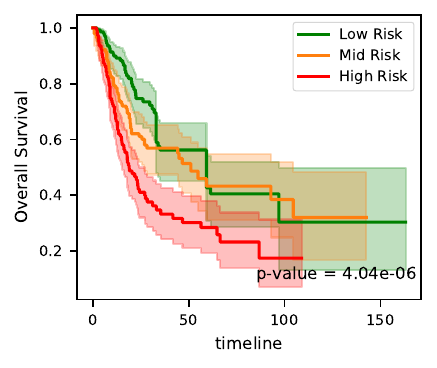}
\end{subfigure}
&
\begin{subfigure}{0.24\textwidth}
    \centering
    \includegraphics[width=\linewidth]{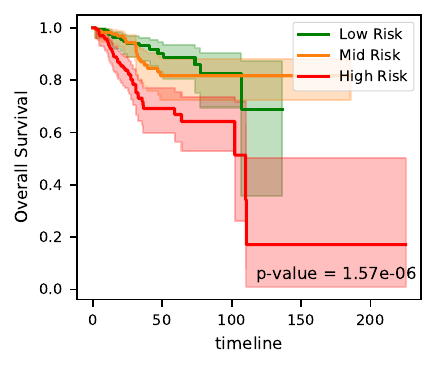}
\end{subfigure}
&
\begin{subfigure}{0.24\textwidth}
    \centering
    \includegraphics[width=\linewidth]{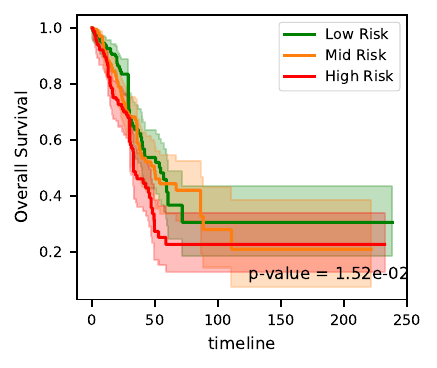}
\end{subfigure}
&
\begin{subfigure}{0.24\textwidth}
    \centering
    \includegraphics[width=\linewidth]{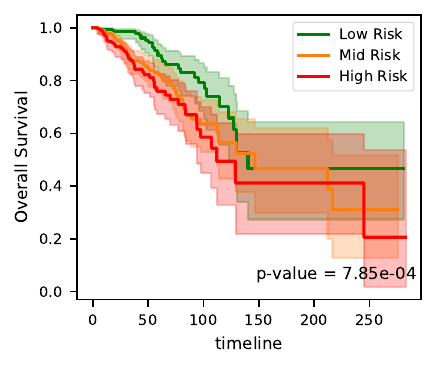}
\end{subfigure}
\\[-0.05cm]

\multicolumn{4}{c}{\textbf{(b) LwF}}
\\[0.35cm]

\begin{subfigure}{0.24\textwidth}
    \centering
    \includegraphics[width=\linewidth]{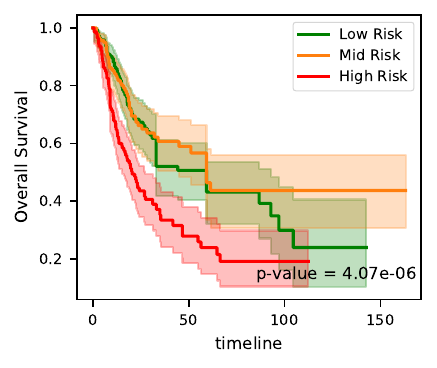}
\end{subfigure}
&
\begin{subfigure}{0.24\textwidth}
    \centering
    \includegraphics[width=\linewidth]{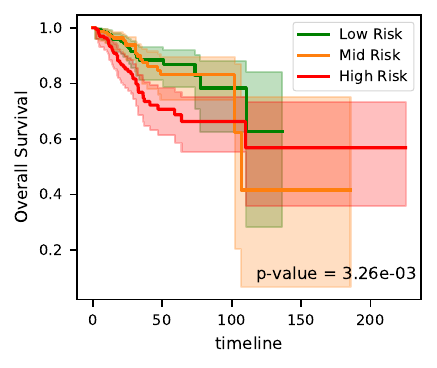}
\end{subfigure}
&
\begin{subfigure}{0.24\textwidth}
    \centering
    \includegraphics[width=\linewidth]{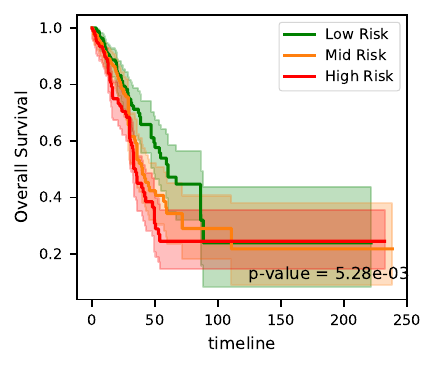}
\end{subfigure}
&
\begin{subfigure}{0.24\textwidth}
    \centering
    \includegraphics[width=\linewidth]{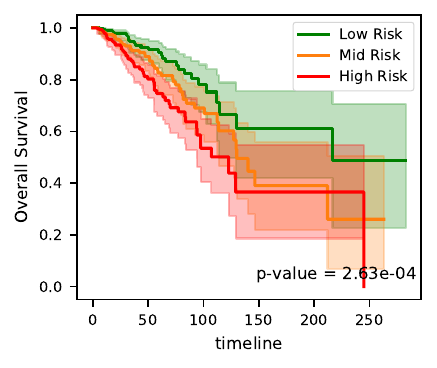}
\end{subfigure}
\\[-0.05cm]

\multicolumn{4}{c}{\textbf{(c) DER}}
\\[0.35cm]

\begin{subfigure}{0.24\textwidth}
    \centering
    \includegraphics[width=\linewidth]{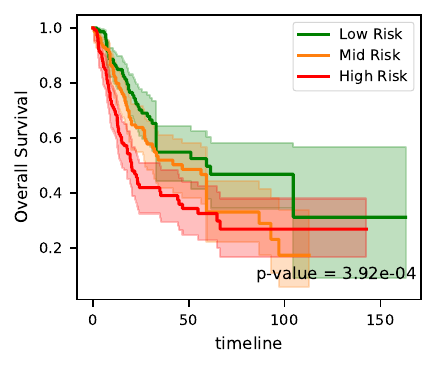}
\end{subfigure}
&
\begin{subfigure}{0.24\textwidth}
    \centering
    \includegraphics[width=\linewidth]{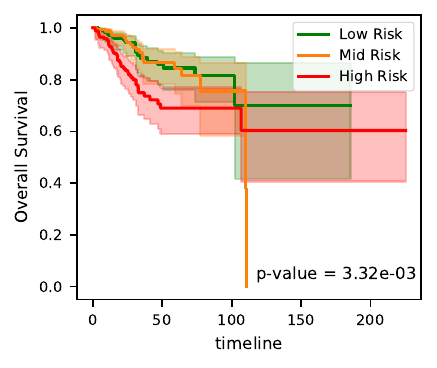}
\end{subfigure}
&
\begin{subfigure}{0.24\textwidth}
    \centering
    \includegraphics[width=\linewidth]{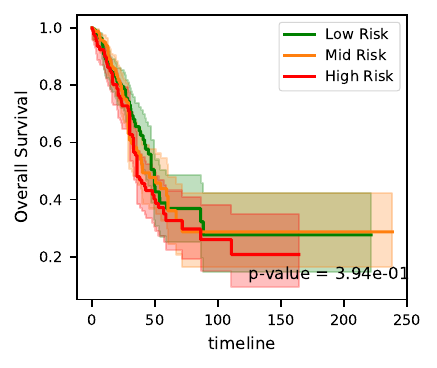}
\end{subfigure}
&
\begin{subfigure}{0.24\textwidth}
    \centering
    \includegraphics[width=\linewidth]{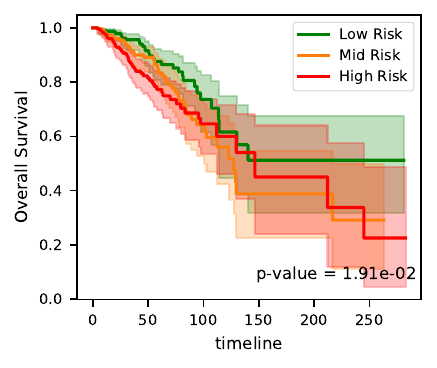}
\end{subfigure}
\\[-0.05cm]

\multicolumn{4}{c}{\textbf{(d) DER++}}
\\[0.35cm]

\begin{subfigure}{0.24\textwidth}
    \centering
    \includegraphics[width=\linewidth]{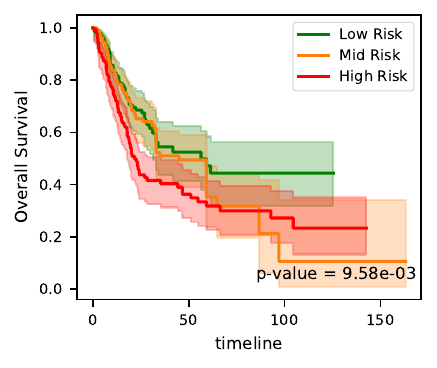}
\end{subfigure}
&
\begin{subfigure}{0.24\textwidth}
    \centering
    \includegraphics[width=\linewidth]{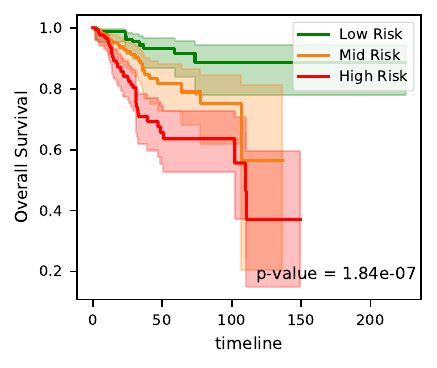}
\end{subfigure}
&
\begin{subfigure}{0.24\textwidth}
    \centering
    \includegraphics[width=\linewidth]{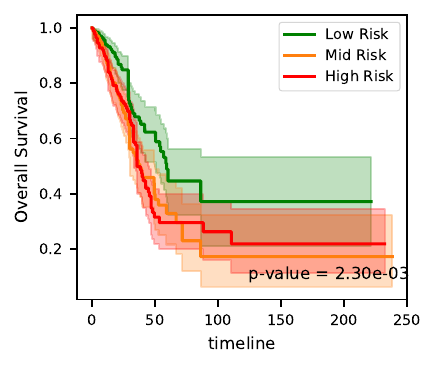}
\end{subfigure}
&
\begin{subfigure}{0.24\textwidth}
    \centering
    \includegraphics[width=\linewidth]{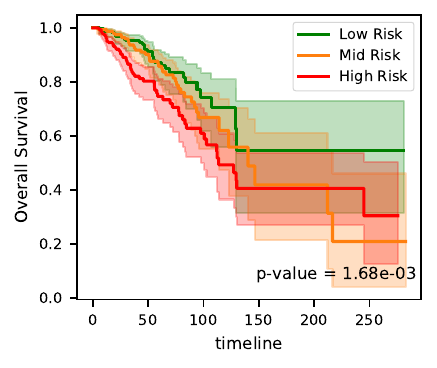}
\end{subfigure}
\\[-0.05cm]

\multicolumn{4}{c}{\textbf{(e) MergeSurv-OFA}}
\\[0.35cm]

\begin{subfigure}{0.24\textwidth}
    \centering
    \includegraphics[width=\linewidth]{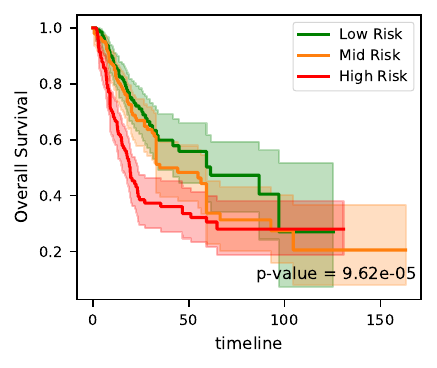}
\end{subfigure}
&
\begin{subfigure}{0.24\textwidth}
    \centering
    \includegraphics[width=\linewidth]{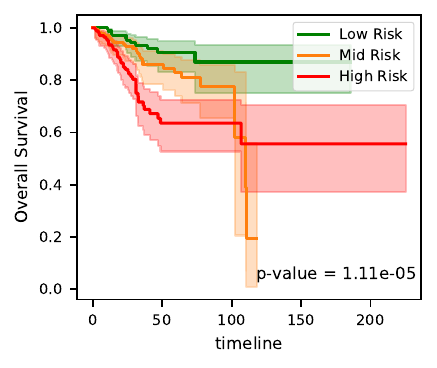}
\end{subfigure}
&
\begin{subfigure}{0.24\textwidth}
    \centering
    \includegraphics[width=\linewidth]{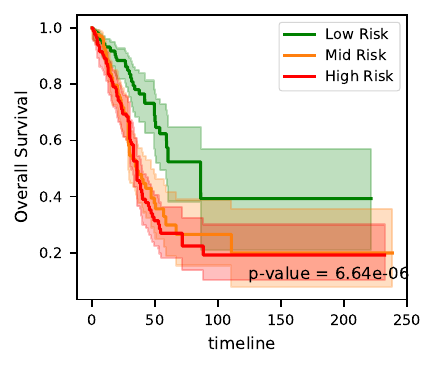}
\end{subfigure}
&
\begin{subfigure}{0.24\textwidth}
    \centering
    \includegraphics[width=\linewidth]{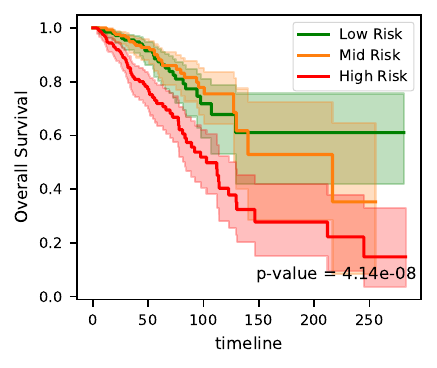}
\end{subfigure}
\\[-0.05cm]

\multicolumn{4}{c}{\textbf{(f) MergeSurv-VEA}}

\end{tabular}

\caption{Kaplan--Meier survival curves of selected continual learning methods across four cancer cohorts (Zooming $4\times$ for viewing ease).}
\label{fig:km_all}
\end{figure}
\clearpage

\noindent The differences are more noticeable for the C-Index IPCW, but the average scores still remain within a relatively narrow range of 0.6170--0.6294. Specifically, Order~\textcircled{1} achieves the highest average C-Index IPCW of 0.6294 and a positive BWT of 0.0011, whereas Order~\textcircled{2} yields the lowest forgetting of 0.0094. Overall, these results suggest that the proposed merging-based framework can stably integrate knowledge across multiple cohorts while remaining relatively robust to changes in task order, making it suitable for practical continual learning scenarios in which the sequence of incoming tasks cannot be predetermined.

\subsection{Kaplan-Meier Analysis}
To provide a more comprehensive evaluation of survival stratification performance, Fig.~\ref{fig:km_all} presents the Kaplan--Meier curves~\cite{kaplan1958nonparametric} for low-, mid-, and high-risk groups, which are defined according to task-wise tertiles of the out-of-fold predicted Cox risk scores for each method. The reported log-rank $p$-values are unadjusted, with no correction for multiple comparisons. On BLCA, most methods fail to maintain consistent risk ordering, where the low-risk curve occasionally falls below the high-risk curve, indicating poor stratification. On UCEC, MergeSurv-OFA shows clearer separation among risk groups, with the high-risk group consistently exhibiting the lowest survival probability. On LUAD, most baselines exhibit substantial overlap between the mid- and high-risk groups, whereas MergeSurv-VEA preserves a more consistent risk ordering. On BRCA, MergeSurv-VEA achieves the clearest separation between risk groups, while other methods produce overlapping survival curves. Furthermore, MergeSurv obtains the lowest log-rank p-values~\cite{mantel1966evaluation} on three out of four datasets, including UCEC ($1.84\times10^{-7}$ with OFA), LUAD ($6.64\times10^{-6}$ with VEA), and BRCA ($4.14\times10^{-8}$ with VEA), suggesting stronger statistical separation between predicted risk groups.

\section{Conclusion}
In this study, we presented MergeSurv, a merging-based continual learning framework for survival analysis on Whole Slide Images (WSIs). Unlike existing rehearsal-based continual learning approaches, MergeSurv avoids storing previous WSIs by independently fine-tuning task-specific models and sequentially merging their parameters into a unified model. We further investigated two inference strategies, namely One-for-All (OFA) and Voting-Expert Aggregation (VEA), for continual survival risk prediction across multiple cancer cohorts. Experimental results on four TCGA datasets demonstrated that MergeSurv effectively mitigates catastrophic forgetting while achieving competitive survival prediction performance compared with regularization-based and replay-based continual learning methods. In particular, MergeSurv-VEA achieved strong risk stratification performance and consistently low forgetting across sequential survival tasks. Overall, the results suggest that model merging is a promising direction for scalable and privacy-preserving continual learning in computational pathology.


%
%
%
\bibliographystyle{splncs04}
\bibliography{refs}

@article{cox1972regression,
  title={Regression Models and Life-Tables},
  author={Cox, David R.},
  journal={Journal of the Royal Statistical Society: Series B (Methodological)},
  volume={34},
  number={2},
  pages={187--202},
  year={1972},
  doi={10.1111/j.2517-6161.1972.tb00899.x}
}

@article{yao2020deepattnmisl,
  title={Whole Slide Images Based Cancer Survival Prediction Using Attention Guided Deep Multiple Instance Learning Networks},
  author={Yao, Jiawen and Zhu, Xinliang and Jonnagaddala, Jitendra and Hawkins, Nicholas and Huang, Junzhou},
  journal={Medical Image Analysis},
  volume={65},
  pages={101789},
  year={2020},
  doi={10.1016/j.media.2020.101789}
}

@article{clam2021,
  title={Data-efficient and weakly supervised computational pathology on whole-slide images},
  author={Lu, Ming Y and Williamson, Drew FK and Chen, Tiffany Y and Chen, Richard J and Barbieri, Matteo and Mahmood, Faisal},
  journal={Nature Biomedical Engineering},
  volume={5},
  number={6},
  pages={555--570},
  year={2021},
  publisher={Nature Publishing Group UK London}
}

@article{wang2024continualsurvey,
  title={A Comprehensive Survey of Continual Learning: Theory, Method and Application},
  author={Wang, Liyuan and Zhang, Xingxing and Su, Hang and Zhu, Jun},
  journal={IEEE Transactions on Pattern Analysis and Machine Intelligence},
  volume={46},
  number={8},
  pages={5362--5383},
  year={2024},
  doi={10.1109/TPAMI.2024.3367329}
}

@inproceedings{bui2026merge,
  author = {Bui, Doanh C. and Ngo, Ba Hung and
            Pham, Hoai Luan and Nguyen, Khang and
            Nguyen, Ma{\"i} K. and Nakashima, Yasuhiko},
  title = {{MergeSlide}: Continual Model Merging and
           Task-to-Class Prompt-Aligned Inference for
           Lifelong Learning on Whole Slide Images},
  booktitle = {Proceedings of the IEEE/CVF Winter Conference
               on Applications of Computer Vision},
  pages = {4859--4868},
  year = {2026},
  doi = {10.1109/WACV61042.2026.00472}
}

@article{yu2026consurv,
  title={ConSurv: Multimodal Continual Learning for Survival Analysis},
  author={Yu, Dianzhi and Xiong, Conghao and Chen, Yankai and Cui, Wenqian and Zhang, Xinni and Zhang, Yifei and Chen, Hao and Sung, Joseph J. Y. and King, Irwin},
  volume={40},
  DOI={10.1609/aaai.v40i33.40013},
  number={33},
  journal={Proceedings of the AAAI Conference on Artificial Intelligence},
  year={2026},
  month={Mar.},
  pages={27899-27907}
}

@article{kirkpatrick2017overcoming,
  title={Overcoming catastrophic forgetting in neural networks},
  author={Kirkpatrick, James and Pascanu, Razvan and Rabinowitz, Neil and Veness, Joel and Desjardins, Guillaume and Rusu, Andrei A and Milan, Kieran and Quan, John and Ramalho, Tiago and Grabska-Barwinska, Agnieszka and others},
  journal={Proceedings of the National Academy of Sciences},
  volume={114},
  number={13},
  pages={3521--3526},
  year={2017},
  publisher={National Academy of Sciences}
}

@article{li2018learning,
  title   = {Learning without Forgetting},
  author  = {Li, Zhizhong and Hoiem, Derek},
  journal = {IEEE Transactions on Pattern Analysis and Machine Intelligence},
  volume  = {40},
  number  = {12},
  pages   = {2935--2947},
  year    = {2018},
  doi     = {10.1109/TPAMI.2017.2773081}
}

@inproceedings{rolnick2019experience,
 author = {Rolnick, David and Ahuja, Arun and Schwarz, Jonathan and Lillicrap, Timothy and Wayne, Gregory},
 booktitle = {Advances in Neural Information Processing Systems},
 pages = {},
 publisher = {Curran Associates, Inc.},
 title = {Experience Replay for Continual Learning},
 volume = {32},
 year = {2019}
}

@article{buzzega2020der,
  title={Dark experience for general continual learning: a strong, simple baseline},
  author={Buzzega, Pietro and Boschini, Matteo and Porrello, Angelo and Abati, Davide and Calderara, Simone},
  journal={Advances in Neural Information Processing Systems},
  volume={33},
  pages={15920--15930},
  year={2020}
}

@inproceedings{lee2025comel,
  author = {Lee, Byung Hyun and Jeong, Wongi and
            Han, Woojae and Lee, Kyoungbun and Chun, Se Young},
  title = {Continual Multiple Instance Learning with
           Enhanced Localization for Histopathological
           Whole Slide Image Analysis},
  booktitle = {Proceedings of the IEEE/CVF International
               Conference on Computer Vision},
  pages = {23232--23242},
  year = {2025},
  doi = {10.1109/ICCV51701.2025.02157}
}

@inproceedings{huang2023conslide,
  title={ConSlide: Asynchronous Hierarchical Interaction Transformer with Breakup-Reorganize Rehearsal for Continual Whole Slide Image Analysis},
  author={Huang, Yanyan and Zhao, Weiqin and Wang, Shujun and Fu, Yu and Jiang, Yuming and Yu, Lequan},
  booktitle={Proceedings of the IEEE/CVF International Conference on Computer Vision},
  pages={21349--21360},
  year={2023}
}

@inproceedings{
bui2025zeroslide,
title={ZeroSlide: Is Zero-Shot Classification Adequate for Lifelong Learning in Whole-Slide Image Analysis in the Era of Pathology Vision-Language Foundation Models?},
author={Doanh C. Bui and Pham Hoai Luan and Vu Trung Duong Le and Vu Tuan Hai and Van Duy Tran and Yasuhiko Nakashima},
booktitle={MICCAI Student Board EMERGE Workshop: Empowering Medical Information Computing and Research through Early-career Guidance and Expertise},
year={2025},
}

@InProceedings{wortsman2022modelsoups,
  title = {Model soups: averaging weights of multiple fine-tuned models improves accuracy without increasing inference time},
  author = {Wortsman, Mitchell and Ilharco, Gabriel and Gadre, Samir Ya and Roelofs, Rebecca and Gontijo-Lopes, Raphael and Morcos, Ari S and Namkoong, Hongseok and Farhadi, Ali and Carmon, Yair and Kornblith, Simon and Schmidt, Ludwig},
  booktitle = {Proceedings of the 39th International Conference on Machine Learning},
  pages = {23965--23998},
  year = {2022},
  volume = 	 {162},
  series = 	 {Proceedings of Machine Learning Research},
  publisher =    {PMLR},
}

@inproceedings{ilharco2023taskarithmetic,
  author = {Ilharco, Gabriel and Ribeiro, Marco Tulio and
            Wortsman, Mitchell and Gururangan, Suchin and
            Schmidt, Ludwig and Hajishirzi, Hannaneh and
            Farhadi, Ali},
  title = {Editing Models with Task Arithmetic},
  booktitle = {International Conference on Learning Representations},
  year = {2023}
}

@inproceedings{tang2025merging,
  author = {Tang, Anke and Yang, Enneng and Shen, Li and
            Luo, Yong and Hu, Han and Zhang, Lefei and
            Du, Bo and Tao, Dacheng},
  title = {Merging on the Fly Without Retraining:
           A Sequential Approach to Scalable Continual Model Merging},
  booktitle = {Advances in Neural Information Processing Systems},
  volume = {38},
  year = {2025}
}

@article{ding2025multimodal,
  author = {Ding, Tong and Wagner, Sophia J. and Song, Andrew H. and
            Chen, Richard J. and Lu, Ming Y. and Zhang, Andrew and
            Vaidya, Anurag J. and Jaume, Guillaume and
            Shaban, Muhammad and others},
  title = {A Multimodal Whole-Slide Foundation Model for Pathology},
  journal = {Nature Medicine},
  volume = {31},
  pages = {3749--3761},
  year = {2025},
  doi = {10.1038/s41591-025-03982-3}
}

@article{harrell1996multivariable,
  title={Multivariable Prognostic Models: Issues in Developing Models, Evaluating Assumptions and Adequacy, and Measuring and Reducing Errors},
  author={Harrell, Frank E. and Lee, Kerry L. and Mark, Daniel B.},
  journal={Statistics in Medicine},
  volume={15},
  number={4},
  pages={361--387},
  year={1996},
}

@article{uno2011cstatistics,
  title={On the C-statistics for Evaluating Overall Adequacy of Risk Prediction Procedures with Censored Survival Data},
  author={Uno, Hajime and Cai, Tianxi and Pencina, Michael J. and D'Agostino, Ralph B. and Wei, L. J.},
  journal={Statistics in Medicine},
  volume={30},
  number={10},
  pages={1105--1117},
  year={2011},
}

@inproceedings{chaudhry2018riemannian,
  title     = {Riemannian Walk for Incremental Learning: Understanding Forgetting and Intransigence},
  author    = {Chaudhry, Arslan and Dokania, Puneet K. and Ajanthan, Thalaiyasingam and Torr, Philip H. S.},
  booktitle = {Proceedings of the European Conference on Computer Vision (ECCV)},
  year      = {2018},
  pages     = {556--572},
  doi       = {10.1007/978-3-030-01252-6_33}
}

@inproceedings{lopez2017gradient,
  title={Gradient Episodic Memory for Continual Learning},
  author={Lopez-Paz, David and Ranzato, Marc'Aurelio},
  booktitle={Advances in Neural Information Processing Systems},
  volume={30},
  year={2017}
}

@article{kaplan1958nonparametric,
  author = {Kaplan, Edward L. and Meier, Paul},
  title = {Nonparametric Estimation from Incomplete Observations},
  journal = {Journal of the American Statistical Association},
  volume = {53},
  number = {282},
  pages = {457--481},
  year = {1958},
  doi = {10.1080/01621459.1958.10501452}
}

@article{mantel1966evaluation,
  author = {Mantel, Nathan},
  title = {Evaluation of Survival Data and Two New Rank Order
           Statistics Arising in Its Consideration},
  journal = {Cancer Chemotherapy Reports},
  volume = {50},
  number = {3},
  pages = {163--170},
  year = {1966}
}
%
\end{document}